# Critical Review for One-class Classification: recent advances and the reality behind them

Toshitaka Hayashi[1] [0000-0002-7599-4404], Dalibor Cimr[1][0000-0003-2197-8553], Hamido Fujita[2,3, *] [0000-0001-5256-210X], Richard Cimler[1] [0000-0001-6712-9894]

[1] Faculty of Science, University of Hradec Kralove, Hradec Kralove, Czech Republic

[2] Malaysia-Japan International Institute of Technology (MJIIT),
Universiti Teknologi Malaysia, Kuala Lumpur, 54100, Malaysia

[3] Regional Research Center, Iwate Prefectural University, Takizawa, Japan

**Abstract.** This paper offers a comprehensive review of one-class classification (OCC), examining the technologies and methodologies employed in its implementation. It delves into various approaches utilized for OCC across diverse data types, such as feature data, image, video, time series, and others. Through a systematic review, this paper synthesizes prominent strategies used in OCC from its inception to its current advancements, with a particular emphasis on the promising application. Moreover, the article criticizes the state-of-the-art (SOTA) image anomaly detection (AD) algorithms dominating one-class experiments. These algorithms include outlier exposure (binary classification) and pretrained model (multi-class classification), conflicting with the fundamental concept of learning from one class. Our investigation reveals that the top nine algorithms for one-class CIFAR10 benchmark are not OCC. We argue that binary/multi-class classification algorithms should not be compared with OCC.

**Keywords:** One-class classification, Anomaly detection, Self-supervised learning,

---
[*]Corresponding author: Professor Hamido Fujita email: <fujitahamido@utm.my>, <HFujita-799@acm.org>,
Postal Address: Kotorizawa, 2-27-5, Morioka, Iwate 020-0104, Japan.
<toshitaka.hayashi@uhk.cz> (Toshitaka Hayashi), <dalibor.cimr@uhk.cz>(Dalibor Cimr), <richard.cimler@uhk.cz> (Richard Cimler)

**Abbreviation list**

OCC one-class classification
AD anomaly detection
ML machine learning
DL deep learning
SOTA state-of-the-art
OCSVM one-class support vector machine
LOF local outlier factor
IF Isolation Forest
GMM Gaussian Mixture Model
AE autoencoder
PU positive-unlabeled

# 1 Introduction

Machine learning (ML) has been applied in various applications [1]. Most applications are supervised learning, which trains the model with well-labeled training data. However, supervised classification has several issues, such as class imbalance and noise. A significant concern arises from the inability of models to classify data into unknown classes that were not present in the training data. Gathering labels for all possible classes is unfeasible due to various reasons, including the rarity of some data, the inherent danger associated with collecting certain types of data (e.g., nuclear explosion, mass shooting, privacy issue), or the fact that some classes do not exist yet (e.g., COVID2030).

One-class classification (OCC) is a special supervised classification problem in which only one class is given in training data. OCC aims to classify data into the target class or other classes. This problem necessitates an unsupervised solution, as the traditional supervised classification model will classify all data into one class. OCC is essential because learning from one class is the most straightforward problem for detecting unseen classes not included in the training stage.

Several OCC algorithms have been proposed in the literature; these approaches use the hypothesis to create an arbitrary score function from one class. These functions can be categorized into likelihood, known as the normal/seen score, and dissimilarity, referred to as the abnormal/outlier/unseen score. Subsequently, the threshold value is decided to classify one class and other classes.

Minter[10] first proposed the concept of OCC in 1975 using the keyword "single-class classification." The topic was the Bayesian network, where negative samples were missing. However, this keyword was unpopular and rephrased to "one-class classification."

According to a Scopus keyword search, Hush published the first conference paper[11] containing the keyword "One-class classifier" in 1989. The content was how to train neural networks with one class. Koch et al.[18] wrote the first journal paper in which OCC was used for application purposes. Subsequently, Moya and Hush [19] published the first journal paper on the OCC algorithm, where the classifier was a neural network. However, neural networks were not successful in the 1990s due to hardware limitations.

In the 2000s, support vector data description (SVDD)[20][22] and one-class support vector machine (OCSVM)[21] were proposed; these methods are based on support vector machines (SVM). Moreover, early OCC algorithms, such as local outlier factor [28], isolation forests [33], and reconstruction-based method [56], were developed at the same time. Meanwhile, neural networks were still underrated and behind the SVM due to their computational complexity.

In the early 2010s, deep learning (DL) gained significant attention and outperformed SVM in supervised image classification. In 2017, Ruff et al.[43] proposed deep OCC for image data. Subsequently, fake unseen [50] and subtask-based approaches [64] have been developed.

OCC is related to several keywords, including anomaly detection (AD) [14], novelty detection[15], outlier detection[16], out-of-distribution detection[17], and intrusion detection[15]. AD aims to detect abnormal data and is frequently used as the representative term for all related keywords. Outlier detection is generally synonymous with AD, as an outlier is a synonym for an anomaly.

Novelty detection aims to recognize data that did not exist in the past. This task can encompass OCC, where all training data possess an "old" label. On the other hand, old data may consist of multiple classes. Out-of-distribution detection, meanwhile, targets the identification of samples that are not in the training data distribution. In this scenario, training data encompass more than one class.

It is essential to highlight that these keywords aim to detect specific data, while the goal of OCC is to learn from one class. Table 1 shows the number of articles indexed with these keywords, the oldest papers, and the years they were published. (The number of papers is based on April 19th, 2024.)

**Table 1** Number of the articles indexed in Scopus.

| Searched keyword | Number | The oldest paper | Year |
|---|---|---|---|
| One-class classification | 2,149 | Hush et al. [12] | 1992 |

| One-class classifier | 912 | Hush [11] | 1989 |
| --- | --- | --- | --- |
| One-class learning | 1712 | Koch et al. [13] | 1994 |
| Single-class classification | 36 | Minter [10] | 1975 |
| Anomaly detection | 37,253 | Corwin et al. [14] | 1970 |
| Novelty detection | 1,888 | Earling [15] | 1976 |
| Outlier detection | 8,920 | Gutmann[16] | 1973 |
| Out-of-distribution detection | 366 | Vyas et al. [17] | 2018 |
| Intrusion detection | 35,235 | Earling [15] | 1976 |

This review paper embarks on a thorough exploration of OCC. Section 2 delves into pertinent research contributions, accentuating key advancements in the field. Section 3 outlines potential approaches to utilizing OCC methods effectively. In Section 4, we illustrate real-world applications of OCC techniques. Section 5 represents a significant contribution by offering critical insights into evaluating state-of-the-art (SOTA) AD algorithms within OCC experiments. Section 6 engages in thoughtful discussions concerning the limitations, challenges, and future possible directions encountered in OCC research. Finally, in Section 7, we synthesize our findings to provide a comprehensive overview of the current landscape of OCC methodologies and their implications.

## 2 Relevant work and main contribution

Several related survey papers have been published [2][3][4][5][6][7][8][9]. According to Scopus, Khan et al. published the first OCC survey in 2014[2]. They classified OCC algorithms into OCSVM or non-OCSVMs. Subsequently, Pimentel et al.[3]

published a review paper on novelty detection with the keyword OCC. They separated the algorithms into probability-based, density-based, reconstruction-based, domain-based, and information-theoretic approaches. At that time, the rise of DL was not significant yet.

Besides, Oliveri [4] discusses OCC in food analysis. Non-target samples are problematic in food science. Trittenbach et al.[5] summarized the review of active learning using one-class classifiers. However, the algorithm is semi-supervised learning, which accesses unlabeled data to find information about other classes that are not OCC.

Recently, Ruff et al.[6], Seiya et al.[7], and Pang et al.[8] provided reviews of OCC. The contents included DL and several new materials, such as self-supervised classification subtasks, outlier exposure, and pretrained models. Moreover, Mašková et al.[9] published the review for set AD, where a single "set data" consists of multiple feature vectors.

Apart from the previous reviews, the contribution of this paper is summarized as follows:

- The novelty is the criticism of state-of-the-art AD algorithms evaluated in a "One-class experiment" setting. This paper revealed that the top nine algorithms in "Anomaly detection on One-class CIFAR-10" were binary/multi-class classification. These algorithms used outlier exposure (binary classification) and a pretrained model from ImageNet (multi-class classification). We argue that binary/multi-class classifiers cannot be SOTA for OCC. Subsequently, this paper advocates removing outlier exposure and pretrained models from comparing OCC algorithms.

- The above circumstance raised the hypothesis that successful OCCs are similar to binary/multi-class classification. Accordingly, this paper discusses the gray zone between OCC and binary/multi-class classification. In particular, positive-unlabeled (PU) learning is not OCC because unlabeled data might belong to other classes. On the other hand, using a pretraining model can be legal as OCC, if the training data is totally artificial.
- The manuscript surveys recent research papers from journals and conference proceedings until 2024. It covers meaningful OCC solutions and individual techniques with their applications and discusses possible future directions in the OCC branch.

## 3  One-class classification approaches

OCC algorithms have been developed for various data types, such as feature (tabular), image, time series, video, 3D, and text. Subsection 3.1 summarizes OCC algorithms for feature data, which is the most general data type because all data types are potentially represented as feature vectors. Subsection 3.2 summarizes OCC algorithms for image, video, time series, and 3D data, which are represented as numerical values. Additionally, subsection 3.3 describes algorithms for text and graph data.

### 3.1  Feature data

Feature data is the most general data type because all other data types could be potentially represented as feature vectors. Early OCC algorithms are proposed for feature data.

Tax et al. proposed support vector data description (SVDD)[20], which computes hypersphere surrounding data distribution. Inside the hypersphere is considered one class, while outside is considered unseen. One-class support vector machine (OCSVM)[21] considers mapping function into feature vector space. Then, the algorithm computes a boundary between mapped samples and origin O, which is supposed to be an outlier. SVDD and OCSVM are equivalents, whereas the Gaussian kernel is used [22]. Subsequently, several improvements have been proposed for OCSVM [23]. Li et al. [24] treated the noises in training data as outliers and improved the overall classification accuracy. Hao [25] introduced a fuzzy function to OCSVM. Choi [26] applied the idea of least square SVM [27] into OCSVM and reduced the time complexity.

Local Outlier Factor (LOF) [28] considers local densities computed by distances from the nearest samples. The hypothesis is that outlier samples have different local densities than their nearest samples. Several related methods are proposed [29]. Local correlation integral considers the quantity of neighbor samples[30]. Cluster-based LOF computes local density using clusters[31]. Average Localized proximity[32] is proposed as an improvement of LOF. The authors pointed out that LOF requires the nearest samples of the nearest samples of the nearest samples to compute the local density score.

Isolation Forest (IF) [33] applies a random tree structure to detect outliers. The hypothesis is that outlier samples are isolated in the tree while normal samples are assigned to the same place. The initial isolation forest utilized a totally random tree, where the branches were random attributes and values. The splits were horizontal or vertical in two-dimensional space. Hariri et al. [34] proposed the Extended Isolation Forest; the idea was to create splits that are not horizontal or vertical. However, the split was still linear. Subsequently, Xu et al.[35] proposed Deep Isolation Forest[35], which create non-linear splits. Moreover, Ryu et al.[36] applied untrained neural networks to detect anomalies, just replacing a random tree with a random neural network.

The Gaussian Mixture Model (GMM)[37] approximates the normal distribution using a Gaussian Mixture. Subsequently, the likelihood for one class is computed, and classification is provided.

The algorithms are then extended to image and time series data. These algorithms are categorized into feature extraction, fake unseen, and self-supervised approaches. The feature extraction approach is not applicable to a feature vector that is already extracted, while fake unseen approaches [38][39] and self-supervised approaches [40] were reimported to feature data. Table 2 summarizes the OCC algorithms mentioned in this subsection.

**Table 2** OCC algorithms for feature data

| Algorithm | Author | Description |
| --- | --- | --- |
| SVDD | Tax et al. [20] | Compute the hypersphere surrounding the data distribution. |
| OCSVM | Schölkopf et al.[21] | Solve SVDD by kernel. |
| OCSVM + noise processing | Li et al. [24] | Ignoring noises in training data. |
| Fuzzy OCSVM | Hao [25] | Introduced fuzzy to OCSVM |
| Least square OCSVM | Choi [26] | Reduced complexity of OCSVM |
| Local Outlier Factor (LOF) | Breunig et al.[28] | Anomalies have different local densities compared to the nearest samples. |
| Average Localized proximity | Lenz et al.[32] | Improving LOF |
| Isolation Forest | Liu et al.[33] | Use a random tree structure. |

| Extended Isolation Forest | Hariri et al. [34] | Linear split with a random tree |
| Deep Isolation Forest | Xu et al.[35] | Non-linear split with random tree |
| Untrained Neural Network | Ryu et al. [36] | Use random neural network |
| Gaussian Mixture model | Figuiredo et al. [37] | Gaussian distribution |
| Kang | Kang [38] | Use some training data as a fake unseen class |
| PBC4OCC | Aguilar et al. [39] | |
| OCFSP | Hayashi et al. [40] | Feature slide prediction subtask |

## 3.2 Image, Time series, Video, 3D

Various algorithms are employed to tackle OCC challenges in image, time series, video, and 3D data, which are all represented numerically. These methods are broadly categorized into feature extraction (subsection 3.2.1), fake unseen (subsection 3.2.2), and subtask-based approaches (subsection 3.2.3).

### 3.2.1 Feature extraction approach

This approach applies feature extraction, which is the technique to represent an arbitrary data type as the feature vector. Subsequently, extracted vectors are processed by OCC algorithms for feature data described in the previous section. The core of this approach lies in how to extract feature vectors. For this purpose, several strategies are proposed for each data type.

In the last ten years, DL-based feature extraction has become popular. DL can automatically extract feature vectors that are effective for specific tasks.

Luff et al.[43] proposed Deep support vector data description (DSVDD), which combined autoencoder (AE)-based feature extraction and SVDD. Moreover, several researchers applied subspecies of AE for feature vectors and claimed novelties. However, these feature vectors did not significantly differ from the original AE. DL can automatically extract effective feature vectors for the target task. In this sense, feature vectors from AEs are specialized for reconstruction and are not helpful for classification. Successful OCC algorithms require feature vectors specialized for classification.

Chen et al. [44] trained a self-supervised classification model and applied a subpart of the network for feature extraction. Subsequently, feature vectors were processed by GMM. In this architecture, feature vectors are specialized for classification, which improves AUC scores significantly. However, classification is done between self-labels. Therefore, there might be a gap between the actual other classes.

Reiss et al. [45][46] applied feature extractors derived from a pre-trained classification model for AD. Since the pretrained models learn from multiple classes, feature vectors for other classes can be represented. This architecture represents the SOTA for "one-class learning" experiments. However, this idea contradicts "learning from one class" but is acceptable in AD.

In the time series data, Mauceri et al. [41] combined dissimilarity-based representations with a one-class nearest neighbor classifier. Hayashi et al. [42] created feature vectors based on clustering results for sliding windows. The advantage is that the model does not require retraining for different signal lengths.

Regarding video data, Wang et al.[47][48] proposed an optical flow orientation feature (capturing movement information for each video frame) and processed the feature vectors using OCSVM. Chriki et al. [49] compared DL-based pretrained features and handcrafted features, where the pre-trained features outperformed the handcrafted ones.

Table 3 Feature extraction processes

| Feature extraction process | What kind of feature |
|---|---|
| Clustering | The feature is unknown because of unsupervised learning. |
| Dissimilarity | Distance from specific points |
| Autoencoder | Feature specialized for reconstruction |
| Self-supervised classification | Feature specialized for classification (from one class) |
| Pretrained model | Feature specialized for classification (from multi-classes, illegal as OCC) |

### 3.2.2 Fake unseen approach

This approach generates fake unseen samples and applies traditional supervised classification between one and fake classes. The core requirements include a fake data generation process and a binary classification algorithm. This study provides an overview of the fake data generation process.

Oza et al.[50] created random Gaussian noises as outliers and trained a convolutional neural network. Yang et al.[51] applied Generative Adversarial Network (GAN) for the data generation process. The main concern with this approach is that training data does not support creating these fake unseen samples, making it challenging to deal with real-world data. Mirzaei et al. [52] applied a diffusion model for the outlier generation process. Subsequently, they applied a pretrained model to classify a normal and a fake unseen class.

Hendrycks et al.[53] proposed outlier exposure, which involved importing other datasets as outliers. Liznerski et al.[54] analyzed the quantity of exposed outlier samples and demonstrated that even a small number of outlier samples suffice to achieve a high AUC score. However, this approach conflicts with OCC as the imported dataset is well-labeled.

In the domain of time series data, Zhu et al.[55] applied adversarial training, where adversarial samples are considered fake outliers. Additionally, Hayashi et al.[135] proposed interpretable synthetic signals for explainable one-class time series classification, where fake signals are created based on editing one-class signals. The idea was to explain the samples classified into unknown classes based on how synthetic signals were created.

However, the algorithms, except for outlier exposure, cannot access other classes. Table 4 summarizes the existing fake data generation process.

**Table 4** Fake unseen data generation process

| Name | How to make fake data | Datatype |
| --- | --- | --- |
| Random Gaussian noise | Random Gaussian | Image |
| OCGAN | Create outlier by GAN | Image |
| Diffusion model | Create outlier by diffusion model | Image |
| Outlier exposure | Import a different dataset as outliers. (Illegal as OCC) | Image |

| Adversarial training | Use the adversarial sample as a fake outlier. | Signal |
| Interpretable synthetic signal | Edit the original signal. | Signal |

### 3.2.3 Subtask-based approach

This approach considers an arbitrary subtask and trains a ML model using only one-class samples. The core hypothesis is that the model error for one class is smaller than its error for other classes. The literature proposes several subtasks categorized into reconstruction, transformation, knowledge distillation, and classification (see Table 5). Reconstruction is the most popular subtask, especially AE [56], a well-known neural network for reconstruction consisting of two networks: an encoder and a decoder. However, the result is not satisfactory as AE possesses strong generalization ability. The researchers have attempted variations of AE [57][58], but the results have not improved significantly.

Other researchers have explored alternative reconstruction subtasks without relying on AE. Lee et al. [59] applied transformers, aiming to encode input by attention mechanisms and decode it through a convolutional decoder. Cai et al. [60] extended the transformation-based reconstruction task in pretrained feature space, operating under the hypothesis that distinguishing normal and abnormal samples is more effective in feature space than in the raw image domain. Notably, this method achieved the highest AUC score in the comparative studies. However, this method utilized a feature extraction with a pretrained model, which might be the main reason for improvement.

The transformation subtask aims to transform data into ideal outputs. Hayashi et al.[61] proposed an image transformation network that aims to transform all images into one image, called the goal image. This architecture outperformed the AE-based subtask due to the model's reduced generalization ability. Interestingly, the results vary, corresponding to the goal images, suggesting a correlation between the subtask difficulty and OCC performance.

Besides, Salehi et al.[63] proposed knowledge distillation subtasks that aim to transfer knowledge from a large model to a small model. Similarly, Bergmann et al.[62] introduced a student-teacher network for AD, where the teacher network was a pre-trained model.

Classification subtask creates a self-labeled dataset and considers multi-class classification between self-labels. The self-labeled dataset is created through an arbitrary process. Golan et al.[64] introduced the classification subtask with geometric transformations. Gidaris et al.[65] offered image rotation classification. Ju et al. [66] highlighted that the classification subtask has an issue due to viewpoint changes and modified the method to normalize the viewpoint of images from one class. Tack et al. [67] provided a classification of shifted instances created by several image transformations, such as cutout, Sobel, noise, blur, perm, rotation, and shift, where the combination of rotation and shift gave the best AUC score. According to the comparison, transformation unrelated to color information exhibited the best result. In the realm of time series data, Blazquez-Garcia et al. [68] proposed the classification of signal multiplication. Moreover, Huang et al.[69] introduced the classification of multiresolution.

**Table 5** Existing subtasks

| Subtask | Goal |
|---|---|
| Reconstruction | Reconstruct input. |
| Transformation | Transform data into ideal output(s). |
| Knowledge distillation<br>Student-teacher | Predict model outputs. |
| Classification | Classify self-labeled dataset. |

## 3.3 Other data types

OCC/AD can be extended to other data types with unique structures. Examples of such data types include text and graph data, elaborated upon in subsections 3.3.1 and 3.3.2, respectively.

### 3.3.1 Text data

Text data encompass words, sentences, paragraphs, or larger bodies of text from various outlets such as books, articles, emails, social media, chats, and websites.

Manevitz et al.[70][71] considered OCC algorithms for documents, extracting features from the text, and applying OCSVM[70] and AE[71]. Peng et al.[72] delved into PU learning with OCSVM, while Halvani et al.[73] proposed authorship verification algorithms.

The general process combines feature extraction and OCC algorithms for feature data. Numerous feature extraction algorithms have been proposed for text data [74][75]. These techniques can be combined with existing OCC algorithms for feature data.

### 3.3.2 Graph data

Graph data consists of nodes and edges, where outlier detection was explored at various levels, including node, linkage, and sub-graph [76].

Xu et al.[77] proposed a structural clustering algorithm for graph architecture and determined node outliers. Kipf et al.[78] proposed variational graph AEs, aiming to represent graphs as feature vectors. Mygdalis et al.[81] and Bandyopadhyay et al. [82][83] considered graph embedding techniques to represent graphs as feature vectors.

Li et al.[79] proposed radar, which applies residual analysis (statistical test) to attribute networks. Peng et al. [80] considered reconstruction a subtask for graphs, while Liu et al. [84] and Xu et al.[85] explored self-supervised classification for graph data.

## 4 Applications

This section delineates the application of the OCC framework. Subsection 4.1 presents software tailored for implementing OCC algorithms. Subsection 4.2 discusses the concept of the one-class ensemble, which aims to enhance OCC algorithms or decompose multi-class classification problems. Finally, subsection 4.3 summarizes real-world applications utilizing OCC methodologies.

## 4.1 Software

OCC algorithms have been implemented in various open packages in Python, which can facilitate applying OCC algorithms to applications. The requirement is to prepare feature vectors for ML input.

Scikit-learn [86] is a well-known ML package with early OCC algorithms, such as OCSVM, LOF, IF, and GMM.

Lenz et al. created fuzzy-rough learn [88], in which OCC algorithms are referred to as data descriptors. This package includes average localized proximity and several OCC algorithms for feature data.

PyOD[87] comprises various AD algorithms for feature data, with a few algorithms tailored for image and time-series data. The package also incorporates various subspecies of early algorithms.

TODS [89] is a package specifically designed for time series outlier detection systems. It aims to detect three types of outliers: point, pattern, and system-wise outliers. TODS supports data processing, time-series processing, feature analysis, and detection algorithms. PyGOD [90] serves as an outlier detection library for graph data encompassing 16 graph outlier detection algorithms. These are feature extraction algorithms, focusing mainly on the backbone).

Furthermore, the Papers With Code webpage [116] summarized the implementation of SOTA algorithms, including DL techniques. It is important to note that the provided source codes do not guarantee the algorithms' reproducibility.

## 4.2 One-class classifier ensemble

The one-class ensemble technique combines multiple one-class classifiers to serve two main objectives: enhancing OCC[91][92][93][94] and improving binary/multi-class classification[95][96] to detect unknown data.

Tax et al.[91] proposed combining one-class classifiers. The goal was to combine multiple OCC algorithms to achieve better OCC performance. Besides, Giacinto et al. [92] extracted feature vectors with three perspectives and trained OCC models for each feature. Subsequently, these models were fused to develop an intrusion detection model.

Krawczyk et al. published several papers on one-class classifier ensembles[93]-[97]. The paper [93] proposed a cluster-based ensemble for OCC, which involves splitting data into clusters and applying OCSVMs for each cluster. Finally, all OCSVM models were ensembled to create an OCC model. Another paper [94] considered dynamic selection between ten OCC classifiers to create an OCC model. Moreover, they proposed the strategy to decompose multi-class classification by one-class classifier ensembles [95] and its dynamic ensemble process [96]. Furthermore, they discussed diversity measures for the ensemble process [97]. Their ensemble strategy was called bagging, which integrates classifiers in parallel ways.

Boosting is an ensemble strategy for integrating classifiers sequentially. Various boosting techniques have been proposed for OCC. Ratsch[98] constructed a boosting framework for OCSVM, while Xing et al. [99] introduced a robust approach with Ada-boosting for OCSVMs. However, only OCSVM was considered the boosting target, although other classifiers can be applicable.

## 4.3 Practical applications

OCC algorithms have been applied to several practical applications.

Abnormal event detection[47][48] aims to distinguish between normal and abnormal events by learning only normal video.

In remote sensing [100][101], the task is to classify land types on Earth. Traditional supervised learning methods had to annotate all labels, where the costs were extensive. Li et al. applied OCC [100] and PU learning [101] to achieve the remote sensing classification without costly annotation.

OCC is also applied for wildfire risk estimation [102]. For this purpose, OCC presents a preferable solution over binary classification, as collecting wildfire data through burning natural habitats is not feasible.

The identification task aims to recognize a specific object, and OCC is a suitable solution due to the prevalence of negative data samples and the absence of future counter-examples. Specific applications include VPN identification[104] and software authorship identification [105].

Bot detection [106] aims to classify users into valid persons or bots related to text analysis. However, the bot can simply copy the human's text or vice versa, complicating the classification task.

Landmine detection [107] presents a critical challenge in safeguarding human life on the battlefields. Their solution involved ground-penetrating radar for data collection and considered landmines one class.

Besides, OCC has been applied in various domains, such as defect prediction [103], damage detection [108][109], and fault detection [110][111]. OCC often outperforms supervised learning in these issues, as training data cannot encompass all possible problems.

In healthcare, OCC has been applied to applications such as fMRI analysis for depression detection [112] and pneumonia screening[113]. These datasets commonly suffer from a class imbalance between normal and diseased patients. OCC offers an advantage by training only healthy patients, thereby mitigating the effects of class imbalance.

In addition, OCC finds applications in food science, including food authentication with food safety[114] and crop disease detection [115], where the data pertains to physical food items, and accessing other classes is challenging. Moreover, collecting anomalous foods might cause problems regarding religions or ethical perspectives.

## 5  Concerns with one-class experiment for anomaly detection.

OCC shares connections with several keywords, such as AD, novelty detection, and out-of-distribution detection. While these terms are technically distinct, with OCC focusing on training a model from one class, the others aim to detect specific problems without requiring the model to learn from only one class.

AD is often discussed synonymously with OCC. All OCC algorithms are AD algorithms that exclusively train on normal samples, while not all AD algorithms belong to the OCC category. The typical examples are outlier exposure and pretrained models.

AD algorithms have often been evaluated using the OCC setting. Outlier exposure and pretrained models outperform OCC algorithms in the "one-class learning" setting because these methods learn binary and multi-classes, respectively. However, this comparison is unfair, and binary or multi-class classifiers should not be considered SOTA for OCC.

In light of this, Table 6 compiles a list of SOTA AD algorithms that conflict with the "one-class learning" concept. For this purpose, the "Anomaly detection on One-class CIFAR10" experiment on the Paper with Code web page [116] is explored (2024/4/19).

The ranked papers were investigated to determine whether the algorithms involved a pretrained model or outlier exposure. A notable exception was where the methodology described involves supervised classification, evidenced by the statement: "**6000 normal and anomaly data images** were used for training MNIST and Fashion-MNIST, and 5000 were used for CIFAR-10" [124].

Table 6 Ranking for "Anomaly detection on One-class CIFAR10" [116]

| Rank | Algorithm | AUROC | One-class classification |
|---|---|---|---|
| 1 | CLIP (OE)[54] | 99.6 | **No (Pretrained model, Outlier exposure)** |
| 2 | Fake It Till You Make It [52] | 99.1 | **No (Pretrained model)** |
| 3 | PANDA-OE [45] | 98.9 | **No (Pretrained model, Outlier exposure)** |
| 4 | Mean-Shifted Contrastive Loss[117] | 98.6 | **No (Pretrained model)** |
| 5 | CLIP (zero-shot) [54] | 98.5 | **No (Pretrained model)** |
| 6 | DINO-FT[46] | 98.4 | **No (Pretrained model)** |
| 7 | Transformaly [118] | 98.3 | **No (Pretrained model)** |
| 8 | CAP [119] | 97.0 | **No (Pretrained model)** |
| 9 | PANDA [45] | 96.2 | **No (Pretrained model)** |
| 10 | CSI [67] | 94.3 | Yes |
| 11 | DUIAD [120] | 92.6 | Yes |
| 12 | DN2 [121] | 92.5 | **No (Pretrained model)** |
| 13 | DisAug CLR [122] | 92.5 | Yes |
| 14 | FCDD [123] | 92 | **No (Outlier Exposure)** |
| 15 | IGD (pre-trained SSL)[44] | 91.25 | Yes |
| 16 | GAN based AD [124] | 90.6 | **No (Supervised classification)** |
| 17 | SSOOD [125] | 90.1 | Yes |
| 18 | SSD [126] | 90.0 | Yes |
| 19 | UTAD [127] | 88.4 | Yes |
| 20 | GOAD [128] | 88.2 | Yes |
| 21 | ARNET[129] | 86.6 | Yes |
| 22 | Reverse Distillation[130] | 86.5 | **No (Pretrained model)** |
| 23 | ADT [64] | 86.0 | Yes |
| 24 | IGD (pretrained) [44] | 83.68 | **No (Pretrained model)** |

| 25 | ESAD [131] | 83.3 | **No (semi-supervised learning)** |

Interestingly, the top nine "one-class CIFAR10" algorithms did not adhere to the OCC paradigm and imported other classes. This conceptual misalignment suggests that recent OCC algorithms had to outperform multi-class classifiers to claim SOTA performance. Future researchers can ignore these techniques when comparing OCC algorithms.

Another remark is that reverse distillation and IGD show low AUC scores despite using pretrained models. Perhaps the models lost representation for multi-classes due to overfitting to one class. In support of this hypothesis, several articles mentioned that the fine-tuning of pretrained models should be limited to a few epochs [54] or a few layers [45].

It is important to note that this criticism is meaningless from an application point of view for AD because the goal is to detect anomalies. A pretrained model learned from multi-classes could be rather beneficial for this purpose. However, one-class experiment results should not be utilized to highlight the advantages of binary or multi-class classifiers. Instead, different experiments should be considered for a fair evaluation.

### 5.1 Binary or multi-classes are better than OCC.

This circumstance suggests that binary/multi-class classifiers outperform OCC. Accessing other classes improves the OCC algorithm's performance, although the algorithm will no longer be OCC. Li et al. reported that PU learning [101] outperforms OCC [100], possibly because PU learning obtains information from unlabeled data belonging to other classes. Moreover, Leevy et al. [134] reported that binary/multi-class classifiers surpass OCC because these techniques can compare classes and compute boundaries between them.

The most successful OCC algorithms use self-supervised classification subtasks. These algorithms create multi-class classification problems from one-class data.

## 5.2    Gray zone OCC

OCC tries to produce binary classification between one class and another by learning only one class of data. Therefore, successful OCC algorithms should be similar to binary/multi-class classification. In this sense, exploring OCC's gray zone could provide hints for improving OCC.

As a core rule, OCC is a classification problem in which only one class is given as training data, and OCC algorithms cannot access **real data** belonging to other classes. This leads us to three fundamental questions:

1)   Is it allowed to access unlabeled real data?

2)   Can we generate fake data from scratch?

3)   Is it possible to generate fake data by transforming training data?

OCC should not access unlabeled data since it could belong to other classes (e.g., vegans cannot eat unlabeled foods because they might be meat!). PU learning is not OCC in this context because the solution finds other classes from unlabeled data. On the other hand, creating fake data from zero is legal because fake data does not belong to other classes. Similarly, creating fake data from training data can be accepted as OCC because the process does not involve other classes.

According to the above discussion, one legal approach is to create a fake dataset. Recently, Kataoka et al. [132] proposed a pretraining strategy utilizing synthetic images generated from formulas, demonstrating the feasibility of training models without natural images. Nevertheless, outlier exposure is illegal as OCC because the method imports other real datasets, whereas importing a fake dataset merely constitutes a fake unseen approach and not an outlier exposure. On the other hand, a pretrained model can be legal if the model learns from an artificial dataset. The main advantage is that the fake dataset can be larger than the real dataset without annotation, which is a promising aspect for OCC.

Besides, creating fake data from training data is permissible as OCC. One can create fake data either by transforming existing training data or by using a further recursive process, such as generating fake data from transformed fake data. In this sense, fake data can diverge significantly from the original training data.

It is important to note that this discussion is political and far from the application perspective. Other researchers do not have to obsess over the OCC concept for application purposes. One straightforward solution is to call all algorithms AD. Nevertheless, binary/multi-class classification algorithms should be removed from the "one-class learning" experiment.

# 6  Discussion

In this section, we comprehensively examine OCC, delving into its advantages and limitations in Subsection 6.1. Here, we dissect the nuanced facets contributing to its efficacy and where it might encounter challenges. Additionally, in Subsection 6.2, we embark on a forward-looking exploration of the future trajectories and potential advancements within the realm of OCC. This section not only elucidates the current landscape but also offers glimpses into the evolving horizons of OCC methodologies and their implications.

## 6.1    Advantages and limitations.

The advantage of OCC is its ability to detect unknown classes not included in the training data, an aspect that binary or multi-class classification cannot cover. Moreover, OCC can avoid class imbalance problems, as imbalance does not happen from one class. Furthermore, OCC tends to incur lower annotation costs than binary/multi-class classification.

The main challenge of OCC is deciding the threshold value between one class and other classes. While most OCC algorithms overlook this aspect and rely on evaluation metrics such as the Area under the ROC curve (AUROC), which is computed without explicitly deciding the threshold value, real-world applications necessitate the determination of these thresholds. Moreover, other classes cannot be involved in determining the threshold value.

## 6.2    Future direction

OCC should obsess over the concept of learning from one class. Outlier exposure and pretrained models should be considered illegal for OCC because these methods learn multi-classes. These methods should be removed from the OCC experiments unless they utilize a completely artificial dataset, potentially falling into a gray area within the OCC framework.

To our knowledge, OCC algorithms for feature data and feature extraction approaches do not have interesting future work. Several combinations of feature extraction methods and OCC algorithms have not yet been explored, so such experiments may not yield compelling insights. Moreover, the future fake unseen approach seems unpromising without outlier exposure or pretrained models.

In contrast, the subtask-based approach is the most promising avenue for pure/legal OCC. Different subtasks hold significant potential, prompting consideration of whether better subtasks exist than self-labeled classification. It is important to note that novel subtask-based OCC might make a similar problem as "novel metaphor-based metaheuristics" in the future, where numerous researchers proposed many optimization algorithms with animal names or other metaphors. They can create new algorithms as the number of metaphors, resulting in a lack of novelty except for their metaphors [133]. Currently, subtask-based OCC remains unaffected by this issue because explored subtasks have been limited yet. However, we do not recommend creating novel metaphor-based subtasks.

Another direction is extending OCC and AD algorithms to other data types, such as time series or video data. These data types are underexplored compared to feature or image data. A simple strategy is to apply image OCC/AD algorithms to other data types.

The future OCC is not promising in real-world applications because several processes have been developed to import multi-classes for AD. Generally, binary/multi-class classification is better than OCC. Nevertheless, OCC could continue to improve the detection of unknown classes (e.g., COVID2030).

# 7 Conclusion

This paper offers a comprehensive review of OCC and AD algorithms, mainly through the lens of OCC experiments. While outlier exposure and pretrained models emerge as SOTA image AD methods, their reliance on binary and multi-class classification contradicts the essence of one-class learning. Consequently, these strategies should be excluded from one-class experiments to maintain methodological fairness.

Conversely, the AD community does not have to obsess over OCC, as the primary goal is to detect anomalies. Practical applications should import other classes to enhance performance. However, the experiments should differ from OCC to ensure a fair and transparent evaluation process.

In summary, while acknowledging the advancements and challenges within the OCC and AD domains, it is imperative to maintain clarity and fairness in experimental design and evaluation methodologies to advance the field effectively.

## CRediT authorship contribution statement

**Toshitaka Hayashi**: Conceptualization, Writing – original draft, Investigation. **Dalibor Cimr:** Writing – original draft. **Hamido Fujita:** Writing – review and editing, Supervision. **Richard Cimler:** Project administration, Funding acquisition.

**Declaration of Competing Interest**

The authors declare that they have no known competing financial interests or personal relationships that could have appeared to influence the work reported in this paper.

**Acknowledgment**

This study is supported by JSPS/Japan KAKENHI (Grants-in-Aid for Scientific Research) #JP-23K11235.

The data described/study is from the project, Research of Excellence on Digital Technologies and Wellbeing CZ.02.01.01/00/22_008/0004583", which is co-supported by the European Union.